\documentclass[letterpaper, 10 pt, conference]{ieeeconf}  

\IEEEoverridecommandlockouts                              

\usepackage{comment}
\usepackage{svg}
\usepackage{bbding}
\usepackage{pifont}
\usepackage{xcolor}
\usepackage{gensymb}
\usepackage{booktabs} 		
\usepackage{hyperref}

\usepackage{xspace}
\makeatletter
\DeclareRobustCommand\onedot{\futurelet\@let@token\@onedot}

\def\@onedot{\ifx\@let@token.\else.\null\fi\xspace}

\def\eg{\emph{e.g}\onedot} 
\def\ie{\emph{i.e}\onedot}

\makeatother
\usepackage{amsmath}

\usepackage{tikz,xcolor,hyperref}
\usepackage{cleveref}

\usepackage{graphicx}
\usepackage{array}

\usepackage{nicematrix}
\usepackage{multirow}
\usepackage{makecell}

\newcommand{\ImgER}[4]{%
  \begin{minipage}[t]{#1\linewidth}\footnotesize
    \includegraphics[width=\linewidth]{#2}\\
    \begin{minipage}[t]{0.52\linewidth}\flushleft $\ E_T$ = #3 cm \vspace{0.2cm}\end{minipage}  \hfill
    \begin{minipage}[t]{0.46\linewidth}\flushright $E_R$ = #4$^\circ \ $\vspace{0.2cm}\end{minipage}
   
  \end{minipage}%
}

\usepackage{algorithm}
\usepackage{algpseudocode}

\usepackage{tikz}
\usetikzlibrary{shapes, arrows.meta, positioning}
\usetikzlibrary{backgrounds}

\tikzstyle{process} = [rectangle, minimum width=1.8cm, minimum height=1cm, text centered, draw=black]
\tikzstyle{arrow} = [thick,->,>=stealth]
\tikzstyle{backarrow} = [ultra thick, ->, >=stealth, blue]

\usetikzlibrary{calc}

\definecolor{lime}{HTML}{A6CE39}
\DeclareRobustCommand{\orcidicon}{
	\begin{tikzpicture}
	\draw[lime, fill=lime] (0,0) 
	circle [radius=0.16] 
	node[white] {{\fontfamily{qag}\selectfont \tiny ID}};
	\draw[white, fill=white] (-0.0625,0.095) 
	circle [radius=0.007];
	\end{tikzpicture}
	\hspace{-2mm}
}
\foreach \x in {A, ..., Z}{\expandafter\xdef\csname orcid\x\endcsname{\noexpand\href{https://orcid.org/\csname orcidauthor\x\endcsname}
			{\noexpand\orcidicon}}
}

\title{\LARGE \bf
NeRF-based Visualization of 3D Cues Supporting Data-Driven Spacecraft Pose Estimation
}

\author{Antoine Legrand\orcidA$^{1,2,4}$, Renaud Detry\orcidB$^{2,3}$ and Christophe De Vleeschouwer\orcidC$^{1}$
\thanks{*This work was supported by Aerospacelab and the Walloon Region through the Win4Doc program.}
\thanks{$^{1}$A. Legrand and C. De Vleeschouwer are with ICTEAM, UCLouvain, Belgium {\tt\small antoine.legrand@uclouvain.be}}
\thanks{$^{2}$A. Legrand and R. Detry are with ESAT, KU Leuven, Belgium}
\thanks{$^{3}$R. Detry is with MECH, KU Leuven, Belgium}%
\thanks{$^{4}$A. Legrand is with Aerospacelab, Belgium}}

\begin{document}

\maketitle
\thispagestyle{empty}
\pagestyle{empty}

\begin{abstract}
On-orbit operations require the estimation of the relative 6D pose, \ie, position and orientation, between a chaser spacecraft and its target. While data-driven spacecraft pose estimation methods have been developed, their adoption in real missions is hampered by the lack of understanding of their decision process. This paper presents a method to visualize the 3D visual cues on which a given pose estimator relies. For this purpose, we train a NeRF-based image generator using the gradients back-propagated through the pose estimation network. This enforces the generator to render the main 3D features exploited by the spacecraft pose estimation network. Experiments demonstrate that our method recovers the relevant 3D cues. Furthermore, they offer additional insights on the relationship between the pose estimation network supervision and its implicit representation of the target spacecraft. 
\end{abstract}

\section{INTRODUCTION}

    On-orbit servicing, \ie, inspecting, refueling, or repairing a spacecraft in flight, is gaining in interest among private companies and public space agencies~\cite{henshaw2014darpa,biesbroek2021clearspace}. It requires Rendezvous and Proximity Operations (RPOs) with a known target spacecraft, \ie,  a chaser spacecraft has to operate close, or even to dock, to a target. While tele-operated operations have been the norm over the past, the community now considers autonomous operations as safer, due to their reduced risks of human or transmission link failures ~\cite{sharma2018pose}. 

    A key feature of autonomous operations is the ability of the chaser spacecraft to perceive its environment and, more specifically, to determine its pose, \ie, position and orientation, with respect to the target spacecraft. Autonomous operations can be cooperative or uncooperative. In the former case, the target spacecraft is equipped with known fiduciary markers or inter-satellite communication links that can be used by the chaser to easily estimate the relative pose. In the latter case, the chaser spacecraft must estimate the target pose solely from its sensors. Although several ones have been considered in the literature~\cite{opromolla2017review}, such as LIDAR systems~\cite{christian2013survey}, infrared~\cite{shi2015uncooperative,rondao2022chinet},  time-of-flight~\cite{martinez2017pose} or stereo~\cite{pesce2017stereovision} cameras, many recent works focused on the use of monocular cameras due to their lower cost, mass, power consumption, or bulkiness~\cite{sharma2018pose}.
   
    Pose estimation methods generally rely on neural networks to predict the relative pose directly from the input image~\cite{kisantal2020satellite, pauly2023survey,park2023satellite}. Indeed, these data-driven approaches are more robust than the hand-crafted ones with regard to adverse illumination conditions. Data-driven spacecraft pose estimation methods usually require a CAD model of the target to generate a large set of images which can be used to train a neural network on ground. The weights of the trained network are then uploaded on the chaser spacecraft.
    
    While data-driven pose estimation techniques are popular, they come with three major challenges. First, training a neural network on synthetic images of the target so that it generalizes to real images, \ie, depicting illumination conditions or textures that differ from the synthetic ones, is difficult~\cite{park2023satellite}. Second, the neural network should be sufficiently lightweight to run on space-grade hardware~\cite{cosmas2020fpga,posso2024real}. While these two challenges have been partially addressed, a third one has surprisingly received little interest. The adoption of data-driven methods for estimating the relative pose of a target in the context of proximity operations is hampered by the lack of explainability that is inherent to data-driven methods, leaving the following question unanswered: "How can we be sure that the network we thoroughly trained on ground will not dramatically fail once in orbit ?" Our work presents a novel method for gaining insights on the key features exploited by spacecraft pose estimation networks, thereby increasing the confidence in the model decisions.
    

    \begin{figure}[b]
        \centering
        \begin{tikzpicture}[
          node distance=0.8cm and 1cm,
          box/.style={draw, rounded corners, minimum width=2cm, minimum height=1cm, align=center, font=\small},
          arrow/.style={-{Stealth}, thick},
          label/.style={font=\small, midway, above},
          belowlabel/.style={font=\small}
        ]
        
        \node[box] (net1) {\Large$G_\Phi$};
        \node[box, right=of net1] (net2) {\Large$P_\Theta$};
        \node[box, right=of net2] (loss) {\large$\mathcal{L_P}$};
        
        \node[above=of net1] (input) {\large ($q,t$)};
        
        \draw[arrow] (input) -- (net1);
        \draw[arrow, label] (net1) -- node[label] {\large$\hat{I}$} (net2);
        \draw[arrow, label] (net2) -- node[label] {\large($\hat{q},\hat{t}$)} (loss);
        \node[font=\small, right=-0.05cm of net1, yshift=-0.2cm] {Visual};
        \node[font=\small, right=0.075cm of net1, yshift=-0.525cm] {Cues};

        \draw[arrow] (input.east) -- ++(3,0) -| (loss.north);
        
        \coordinate (start) at ($(loss.south) + (0.85,-0.3)$);
        \coordinate (end) at ($(net1.south) + (-0.6,-0.3)$);
        
        \draw[thick] (start) -- (end);
        \draw[thick] ($(start)+(0,-0.15)$) -- ($(end)+(0,-0.15)$);
        \draw[thick] (start) -- ($(start)+(0,-0.15)$);
        
        \draw[thick, -{Stealth[length=4mm]}] ($(end)+(-0.3,-0.075)$) -- ++(-0.01,0);
        
        \node[belowlabel] at ($(start)!0.5!(end)-(0,0.45)$) {Back-propagation};
        
        \node at (3.7,0.26) {\includegraphics[width=0.42cm]{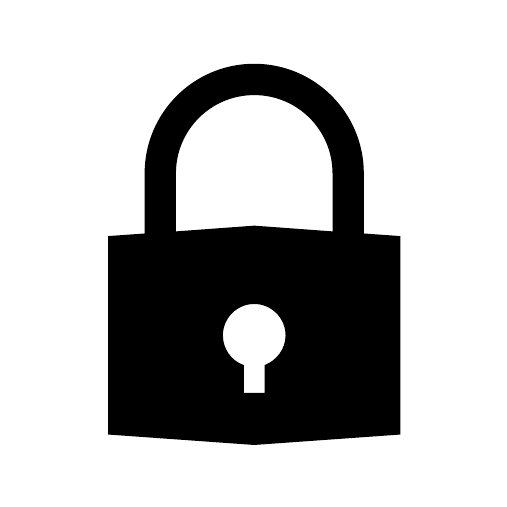}};
        
        \end{tikzpicture}
        \caption{To visualize the 3D cues exploited by a spacecraft pose estimation network $P_{\Theta}$, our method relies on an image generator $G_{\Phi}$ which takes as input a 6D pose and outputs an image. By back-propagating the difference between the pose predicted on that image (by the frozen pose estimator) and the input pose, the generator is trained. Once it has been trained, the generator can synthesize images containing the 3D cues on which the pose estimation network primarily relies.}
        \label{fig_overview}
    \end{figure}
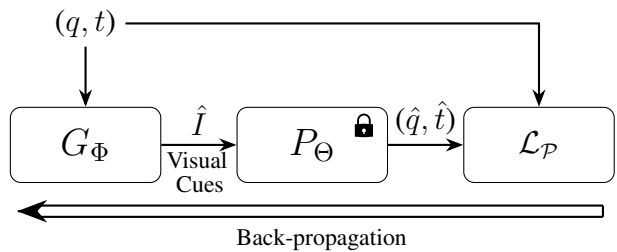

    \Cref{fig_overview} depicts an overview of our method. Our goal is to train an image generator $G_{\Phi}$ that, given a pose label, \ie, orientation $q$, and position $t$, outputs an image $\hat{I}$, which, when given to a trained pose estimator, $P_{\Theta}$, results in an estimated pose $(\hat{q},\hat{t})$ close to the original pose label. The underlying assumption behind our method is that if such a generator $G_{\Phi}$ exists and is able to generate images from any viewpoint that are correctly estimated by the pose estimator, then this generator should have learned to synthesize the target features that are primarily exploited by the pose estimation  network.

    Our contributions are therefore two-fold. First, we present a novel method for visualizing the key target features exploited by a spacecraft pose estimation network. Second, we conduct experiments with our 3D cues visualization method to gain insights on the features exploited by State-of-the-Art spacecraft pose estimation networks.

\section{RELATED WORKS}
\label{sec_related_works}

    The task of estimating the pose, \ie, orientation and position, of an uncooperative, known, target spacecraft relative to a monocular camera has recently attracted the interest of the community. Although numerous neural network architectures have been proposed for this task, they can be classified in three main families. First, keypoint-based methods~\cite{chen2019satellite,cassinis2022ground} rely on the detection of pre-defined keypoints in the image space to compute the pose through a Perspective-n-Point solver~\cite{lepetit2009ep}. Second, end-to-end methods~\cite{sharma2018pose,proencca2020deep,posso2022mobile} directly predict the target pose from the input image through neural networks. They usually rely on a backbone network followed by regression and/or classification heads. Third, hybrid approaches~\cite{legrand2022end,park2024robust,legrand2024domain} combine both the end-to-end approach with the prediction of pre-defined keypoints. Despite their architectural differences, every data-driven spacecraft pose estimation network can be considered as a neural network black box that takes an input image $I$ and outputs a pose $(q,t)$. Furthermore, they all aim at estimating the pose of a specific target spacecraft so that the pose estimation network is expected to overfit the relevant target features.

    Despite a strong interest among the computer vision community, explaining the predictions of a neural network remains an open question. Explainability methods aim at determining which input pixels have the strongest influence on the prediction of a trained neural network. Although several methods exist, they can be classified in two main categories. On one hand, class activation maps~\cite{selvaraju2017grad,smilkov2017smoothgrad,sundararajan2017axiomatic} reflect the derivatives of the network predictions with respect to the input pixels. On the other hand, perturbation-based methods~\cite{ribeiro2016should,petsiuk2018rise} consist in modifying the input image and observing how this perturbation has affected the network prediction. These methods have been developed for both convolutional neural networks~\cite{englebert2024poly}, as well as transformer-based architectures~\cite{englebert2022backward} with a strong emphasis on classification tasks. Unlike those works, our method does not aim at identifying which input pixels contributed to the network decision for a particular input image. Instead, it aims at synthesizing a 3D scene whose rendering in 2D is consistent with the pose estimation model, meaning that the pose predicted from a projected image corresponds to the actual viewpoint parameters used to project the scene. 

\begin{figure*}[t]
  \centering
  \vspace{0.3cm}
  \begin{minipage}[c]{0.45\textwidth}
    \begin{tikzpicture}[node distance=0.8cm, >=Stealth, align=left]

    \node (A) [rectangle, draw] {Pose};
    \node (B) [rectangle, draw, below=of A] {{\small $W \!\times\! H$} rays};
    \node (C) [rectangle, draw, below=of B] {{\small$W \!\times\! H \!\times\! N$} samples};
    \node (D) [rectangle, draw, below=of C] {{\small$W \!\times \! H \! \times \! N$} points};
    \node (E) [rectangle, draw, below=of D] {Image};
    
    \def\rectwidth{3.2}
    \def\rectheight{4.6}
    
    \begin{scope}[on background layer]
        \draw[rounded corners=5pt, fill=red!05, draw=none]
        ($(C) + (-\rectwidth/2, -\rectheight/2)$) 
        rectangle 
        ($(C) + (\rectwidth/2, \rectheight/2)$);
    \end{scope}
    
    \node (tempf) [left=0.8cm of B] {};
    \node (F) at ($(tempf)+(0.65cm,0.55cm)$) [text=red!90, font=\Large\bfseries] {$G_\Phi$};

    \def\distbox{1.5cm}
    
    \draw[->] (A) -- (B) node[very near start, right, text width=7.25cm, anchor=north west] {
        \textbf{$\mathcal{K}$:} Projecting rays through {\small$W \!\times\! H$} image pixels\\\vspace{0.075cm}
        $ \hspace{\distbox} \displaystyle (q,t) \xrightarrow{\mathcal{K}} \left\{ \left(c,d\right)_{ij}\right\}$
    };
    \draw[->] (B) -- (C) node[very near start, right, text width=7.25cm, anchor=north west]{
        \textbf{$\mathcal{S}$:} Sampling N points along each ray\\\vspace{0.075cm}
        $ \hspace{\distbox} \displaystyle \left(c,d\right)_{ij} \xrightarrow{\mathcal{S}} \left\{ \left(x,y,z,\theta,\phi\right)_{ijk}\right\}$
    };
    \draw[->] (C) -- (D) node[very near start, right, text width=7.25cm, anchor=north west]{
        \textcolor{blue!45}{{\textbf{$\mathcal{F}$}}: Neural field inference}\\\vspace{0.075cm}
        $ \hspace{\distbox} \displaystyle \left(x,y,z,\theta,\phi\right)_{ijk} \xrightarrow{\textcolor{blue!45}{\mathcal{F}}} \left(r,g,b,\sigma\right)_{ijk}$
    };
    \draw[->] (D) -- (E) node[very near start, right, text width=7.25cm, anchor=north west] {
        \textbf{$\mathcal{R}$:} Differentiable ray tracing\\ \vspace{0.075cm}
        $ \hspace{\distbox} \displaystyle \left\{ \left(r,g,b,\sigma\right)_{ijk}\right\} \xrightarrow{\mathcal{R}} I $
    };
    
    \end{tikzpicture}
  \end{minipage}
  \hfill
  \begin{minipage}[c]{0.54\textwidth}
    \centering
    \vspace{0.5cm}
    \begin{tikzpicture}[node distance=1cm and 1cm, auto, >=Stealth]
        \tikzstyle{block} = [draw, fill=blue!05, rounded corners, minimum height=3.5cm, minimum width=6.5cm]
        \tikzstyle{component} = [draw, fill=white, rounded corners, minimum height=1cm, minimum width=2.3cm]
        \tikzstyle{input} = [coordinate]
        \tikzstyle{output} = [coordinate]
        \tikzstyle{arrow} = [->, thick]
        
        \node[block] (main) {};   
        \node[input, above=of main, yshift=-0.7cm, xshift=0.35cm] (F) {};  
        \node[input, left=of main, xshift=0.75cm, yshift=1cm] (A) {};
        \node[input, left=of main, xshift=0.75cm, yshift=-1cm] (B) {};
        
        \node[output, right=of main, xshift=-0.6cm, yshift=1cm] (C) {};
        \node[output, right=of main, xshift=-0.6cm, yshift=-1cm] (D) {};
        
        \node[component, right=0.75cm of A] (comp1) {Pos. Encoding};
        \node[component, right=0.75cm of B, fill=gray!10] (comp2) {Dir. Encoding};
        \node[component, right=1cm of comp1] (comp3) {Density Field};
        \node[component, below=1.0cm of comp3] (comp4) {Color Field};
        
        \node[left=0cm of F] {{\textcolor{blue!45}{\Large $\mathcal{F}$}}};
        \node[left=0cm of A] {($x$,$y$,$z$)};
        \node[left=0cm of B] {($\theta$,$\phi$)};
        \node[right=0cm of C] {$\sigma$};
        \node[right=0cm of D] {($r$,$g$,$b$)};
        
        \node (tempa) [right=0.3cm of comp1] {};
        \node (a) at ($(tempa)+(0cm,-0.35cm)$) {$F_{\textnormal{pos}}$};
        \node (tempb) [right=0.3cm of comp2] {};
        \node (b) at ($(tempb)+(0cm,0.35cm)$) {$F_{\textnormal{dir}}$};
        \node (tempc) [below=0.3cm of comp3] {};
        \node (b) at ($(tempc)+(0.25cm,0cm)$) {$F_{\sigma}$};
        
        \draw[arrow] (A) -- (comp1);
        \draw[arrow] (B) -- (comp2);
        \draw[arrow] (comp1) -- (comp3);
        \draw[arrow] (comp3) -- (C);
        \draw[arrow] (comp2) -- (comp4);
        \draw[arrow] (comp3) -- ++(0,-0.5) -| (comp4);
        \draw[arrow] (comp4) -- (D);
    \end{tikzpicture}

  \end{minipage}    
  \caption{\textbf{(Left:)} NeRF rendering pipeline. (i): Rays, denoted by their origin $c$ and unit direction $d$, are projected through every pixel ($i$,$j$) of a calibrated camera of relative pose ($q$,$t$). (ii) $N$ points, \ie, a point being defined by a 3D position ($x$,$y$,$z$) and two viewing angles ($\theta$, $\phi$), are sampled along each ray. (iii) For each point, its color ($r$,$g$,$b$) and density ($\sigma$) are predicted by the neural field $\mathcal{F}$. (iv) For each ray, the value of the corresponding pixel is determined through differentiable ray-tracing techniques. \textbf{(Right)}: Architecture of the neural field $\mathcal{F}$. The input position and viewing angles of a point are mapped to position and direction features, $F_{\textnormal{pos}}$ and $F_{\textnormal{dir}}$, respectively through a learnable 3D grid~\cite{fridovich2023k} and spherical harmonics. The position features are fed in a MLP which approximates the density field to output the density $\sigma$ of the point along density features $F_\sigma$. Density and direction features are processed by a second MLP to predict the color of the point.\vspace{-0.2cm}}
\label{fig_nerf_architecture}
\end{figure*}
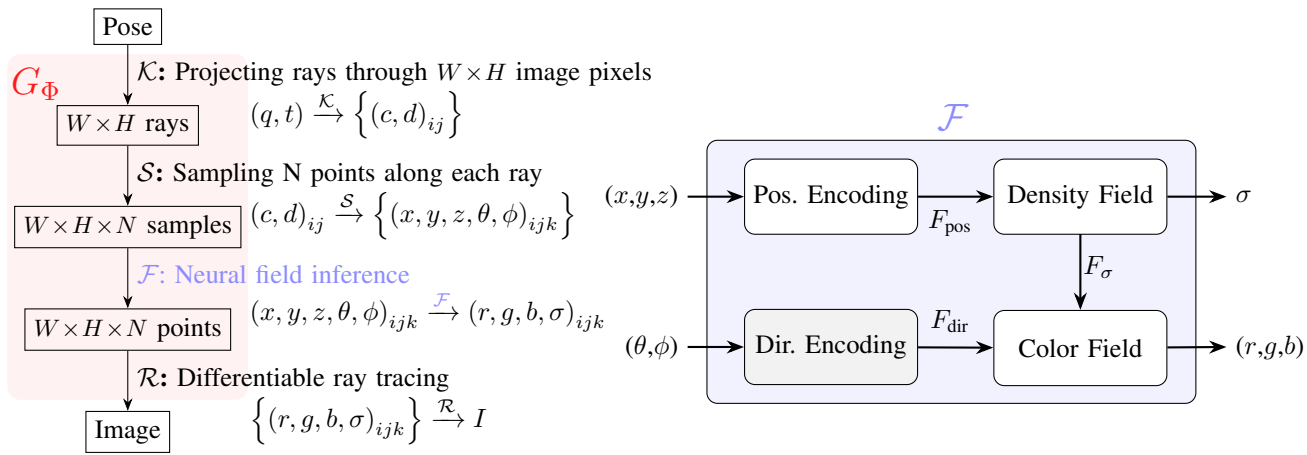

    In this work, the image generator consists in a Neural Radiance Field~\cite{mildenhall2021nerf} (NeRF). A NeRF aims at learning the geometry and appearance of a scene from a set of a few training images along their pose label in order to render novel views of that scene from unseen viewpoints. As depicted in ~\cref{fig_nerf_architecture}, to synthesize an image taken from a given viewpoint, a NeRF projects rays across every pixel of that image. It then samples points along the rays and feed them in a multilayer perceptron (MLP) which outputs their colors and densities. Finally, through differentiable ray-tracing techniques, the color of each pixel is computed from the color and density of all the points along the corresponding rays. The NeRF is iteratively trained on the training images by back-propagating the photometric loss between the training image and the image generated by the NeRF under the corresponding pose label. Although several works~\cite{legrand2024leveraging,legrand2024domain_nerf} explored the use of NeRFs in the context of spacecraft pose estimation, they only considered the Neural Radiance Field as a tool for data augmentation. In this paper, their ability of learning a 3D scene through back-propagation is exploited to visualize the features that support  the predictions of a spacecraft pose estimation network.

\section{METHOD}

This section describes our method for visualizing the features that are  primarily exploited by a spacecraft pose estimation network $P_{\Theta}$. 
As depicted in \Cref{fig_overview}, we rely on a generation network $G_{\Phi}$ which takes as input a 6D pose, \ie, orientation $q$ and position $t$, and outputs an image $\hat{I}$ of resolution $W$x$H$.
Our goal is to train this generator so that the images that it renders contain the features primarily exploited by the pose estimation network $P_{\Theta}$. To determine the generator weights $\Phi^*$ that achieve that goal, we iteratively update them by back-propagating the difference between the input pose and the pose predicted by the frozen pose estimation network $(\hat{q},\hat{t}) = P_{\Theta}(\hat{I}))$, \ie, 
\begin{equation}
    \Phi^* = \arg\min_{\Phi} \mathcal{L_{P}}\left(\Phi; \left(\hat{q},\hat{t}\right), \left(q,t\right)\right)
\end{equation}
where $\mathcal{L_{P}}$ computes the distance between the input and predicted poses in the $\mathrm{SE}(3)$ space.

Although the proposed feature visualization approach is described in general terms, training an image generation network under this formulation may lead to convergence issues. To increase the probability that the training of the generator converges, we rely on an NeRF-based image generator, which is described in \Cref{sec_meth_gen_arch}. \Cref{sec_3d_consistency} introduces the training procedure required to encourage the image generation network to learn 3D-consistent features, and therefore improve its probability of converging.

\subsection{Image Generator Architecture}
\label{sec_meth_gen_arch}
    Our approach could rely on any generation network that takes as input a 6D pose and outputs an image, \eg, pose-conditioned GANs (Generative Adversarial Networks)~\cite{ma2017pose,nguyen2019hologan}. However, taking inspiration from previous works on NeRF-based pose-conditioned image generation~\cite{schwarz2020graf,chan2021pi,chan2022efficient}, our method relies on a Neural Radiance Field~\cite{mildenhall2021nerf}. NeRFs were initially designed to render novel views of a scene, given a set of training images depicting that scene. Here, the NeRF has to learn the 3D features exploited by a pose estimation network through the gradients back-propagated from that network.
    
    \Cref{fig_nerf_architecture} depicts the architecture of our NeRF-based image generator, following $K$-planes~\cite{fridovich2023k}. Given the input pose, rays are projected from the camera focal point through each pixel. For each ray, N points are sampled along the ray through a coarse density model of the scene. For each point, position and direction features are computed from its 3D position and viewing angles. The position features are fed in a first multilayer perceptron (MLP) which predicts density features along the point density. These density features, concatenated to the direction ones, are given to a second MLP to infer the color of the point. Finally, through ray-tracing techniques, the value of each pixel of the image is computed from the densities and colors of the points sampled on the corresponding ray.

    Using a Neural Radiance field as image generator offers several benefits. First, the NeRF representation is inherently \textbf{3D consistent} due to its architecture which relies on rays projected through the scene and the decoupling of the color field from the density one. Second, due to recent architectural improvements, such as the learned positional encoding~\cite{muller2022instant,fridovich2023k}, NeRFs are \textbf{computationally efficient} and can therefore be trained even on a consumer-grade GPU. Third, NeRFs are \textbf{modular}. Hence, to increase the convergence probability, we first train a NeRF on the synthetic dataset on which the pose estimator was trained. Then, we initialize a novel instance of that NeRF from scratch, except for the sampler and the learned positional encoding, which are copied from the pre-trained model and are not updated during the generator optimization. These three architectural characteristics significantly alleviate the convergence issues related to the problem formulation. 
    
\subsection{Ensuring 3D Perception Through Gradient Accumulation}
\label{sec_3d_consistency}

    Even if the convergence issues are partially solved through the image generator architecture, convergence is not guaranteed because of a key difference in the training strategies of NeRFs and pose estimators. Indeed, to ensure 3D consistency and promote convergence, each update of the NeRF weights should be derived from gradients computed across multiple viewpoints. 
    When learned from a set of views, NeRFs can be trained with batches composed of pixels randomly sampled across all input images, effectively mixing viewpoints and therefore promoting 3D consistency. 
    In contrast, pose estimators operate on entire images. Each update to the network’s weights typically requires data from at least one full image, and ideally from multiple training images, although this is not strictly necessary. This constraint is at the heart of the difficulty to train a NeRF with a pose estimation loss.

    Ideally, training a NeRF with a pose estimation loss should involve batches of full images corresponding to multiple viewpoints to ensure 3D perception. However, this would imply tremendous GPU memory requirements. We reduce these requirements by relying on two complementary mechanisms. First, to reduce the per-image generation complexity, we render images at half the resolution expected by the pose estimator and upsample them afterwards. Furthermore, we generate only the pixels that actually correspond to the target spacecraft. The remaining ones, corresponding to the outer space, are set to black. Second, to mimic larger batch sizes containing images taken from multiple viewpoints, we perform gradient accumulation.

    \Cref{algo_training_grad_acc} describes the gradient accumulation strategy required to train the image generation network on a large effective batch size, thereby promoting the 3D consistency of the learned 3D cues. Each step of the algorithm corresponds to the processing of a single full image. First, an image $\hat{I}_s$ is rendered by the image generator $G_\Phi$ from a pose label $(q,t)_s$ sampled in the training dataset $\mathcal{D}$. Then, a loss $\mathcal{L}_s$ is computed between the prediction of the pose estimator $F_\Theta$ on that image and the pose label. Finally, the gradients of the image generator's weights are computed through back-propagation and accumulated. Every $N$ steps, the image generator weights $\Phi$ are updated by the optimizer $O$ according to the average of the stored gradients.
    
    \begin{algorithm}
    \caption{Training $G_{\Phi}$ Through Gradient Accumulation}
    \begin{algorithmic}[1]
    \label{algo_training_grad_acc}
    \Require Image generator $G_{\Phi}$, pose estimation network $P_{\Theta}$, optimizer $O$, loss $\mathcal{L}$, dataset $D$, accumulation steps $N$, total steps $T$
    \State Initialize model parameters $\Phi$
    \State Initialize gradient accumulator $G \gets 0$
    \For{step $s = 1$ to $T$}
        \State Sample pose $(q, t)_s$ from dataset $D$
        \State Generate image $\hat{I}_s \gets G_{\Phi}((q, t)_s)$
        \State Compute loss: $L_s \gets \mathcal{L}(P_{\Theta}(\hat{I}_s), (q, t)_s)$
        \State Compute gradients: $g_s \gets \nabla_\Phi L_s$
        \State Accumulate gradients: $G \gets G + g_s$
        \If{$s \mod N = N-1$}
            \State Update parameters: $\Phi \gets \Phi - O(G / N)$
            \State Reset accumulator: $G \gets 0$
        \EndIf
    \EndFor

    \end{algorithmic}
    \end{algorithm}

\section{EXPERIMENTS}

This section presents a set of experiments conducted with our feature visualization method on a spacecraft pose estimation network. \Cref{sec_val_setup} describes our validation setup, \ie, architecture of the image generation and pose estimation networks and the generator optimization strategy. \Cref{sec_val_main} demonstrates that our method recovers the 3D features that are primarily exploited by a pose estimation network and are sufficient to predict the correct relative pose. \Cref{sec_val_insights} exploits our visualization method to gain some insights on the learning mechanisms of spacecraft pose estimation networks.

\subsection{Validation setup}
\label{sec_val_setup}

    Our experiments are conducted on a pose estimator trained on SPEED+~\cite{park2022speed+}. SPEED+ contains 59,960 synthetic images depicting a spacecraft, TANGO from the PRISMA mission~\cite{gill2007autonomous}, used to train a pose estimation network and assess its accuracy on Hardware-In-the-Loop (HIL) images.
    
    The pose estimation network $P_\Theta$ considered in our experiments is \textbf{SPNv2}~\cite{park2024robust}, \textbf{a state-of-the-art spacecraft pose estimation network} designed to achieve a good accuracy on real images, despite their differences with the synthetic ones. To generalize to unseen domains, SPNv2 relies on strong data augmentation policies applied to a hybrid architecture which regresses the 6D pose and predicts the position of $K$ pre-defined keypoints through heatmaps. (see \cref{sec_related_works}). Its architecture consists in an EfficientNet backbone~\cite{tan2019efficientnet} that predicts feature maps which are then shared by 3 heads. The first head regressed $K=11$ heatmaps that highlight pre-defined keypoints on the spacecraft, \ie, the 8 corners of the spacecraft body and the extremities of its three main antennas. Given the coordinates of these $K$ keypoints, the relative pose can be computed by a Perspective-n-Point solver~\cite{lepetit2009ep}. The second head consists in an EfficientPose~\cite{bukschat2020efficientpose} which regresses the relative pose and bounding box. Finally, the last head performs foreground segmentation. Unless otherwise mentioned, SPNv2 is trained using the same parameters as in the original paper~\cite{park2024robust}.

    The image generation network $G_\Phi$ consists in a NeRF~\cite{mildenhall2021nerf} that follows the $K$-planes implementation~\cite{fridovich2023k} (see \Cref{fig_nerf_architecture}) and is trained using our method. First, we train a NeRF on 200 synthetic images from SPEED+, resized to 256x384 pixels. Then, we initialize a new NeRF and update the sampler weights and the learned positional encoding with the corresponding pretrained weights. The weights of the sampler and positional encoding are frozen during training, only the density and color fields are trained. For each forward pass, only the foreground pixels are given to the network to reduce the GPU memory footprint. A full image of 256x384 pixels is assembled from the foreground ones, the background ones being set to black. The image is then upsampled by a factor 2 to achieve the resolution used during the pose estimation network training, \ie, 512x768 pixels. The image is then normalized as expected by SPNv2 and fed into the pose estimation network which outputs $K$ heatmaps and a pose. 
    
    The loss back-propagated through $P_\Theta$ and $G_\Phi$ consists in the sum of two losses associated to the heatmaps and pose, both of them weighted by a factor $\beta$. This factor, determined through empirical evaluation, is set to 0.01. Indeed, experiments demonstrated that a factor of 0.01 or lower than this value leads to a probability of convergence of approximately 90 \%. The first term is the L2-error between the predicted and target heatmaps while the second term is the SPEED loss~\cite{park2024robust} which combines the angular error with the normalized translation error. The weights of $G_\Phi$ are updated for $T/N$ = 1,000 optimization steps, \ie, each optimization step corresponding to $N$=10 images, while the weights of $P_\Theta$ are frozen. The optimization takes place on a single NVIDIA L40S and lasts about 30 minutes.

    In the following sections, the the angular error ($E_R$, in degrees) and the translation errors ($E_T$, in meters) are computed on the prediction inferred by the pose estimation network on the corresponding image.

\renewcommand{\arraystretch}{4.6}
\begin{table*}[t]
\vspace{0.5cm}
\centering
\caption{\small\label{tab_validation}\textbf{(Top):} Synthetic images from SPEED+~\cite{park2022speed+} and the corresponding error metrics achieved by a pose estimator $P_{\Theta}$. \textbf{(Middle-Bottom):} 3D cues rendered by a NeRF $G_{\Phi}$ trained with our method, under the same viewpoints, along with the metrics achieved by $P_{\Theta}$, considering two training configurations for $G_{\Phi}$, either the neural field $\mathcal{F}$ of $G_{\Phi}$ is trained from scratch  \textbf{(Middle)} or the positional encoding of $\mathcal{F}$ is copied from the pretrained model and only the fields parameters are trained \textbf{(Bottom)}. In both cases, the learned visual cues are sufficient to enable the pose estimator to accurately predict the pose.}
\begin{NiceTabular}{@{}m{1cm}@{} m{1.7cm}@{}*{4}{c@{}}}[hvlines]
\Block{2-2}{\small Synthetic images \\from SPEED+} & & 
    \Block{2-1}{\ImgER{0.205}{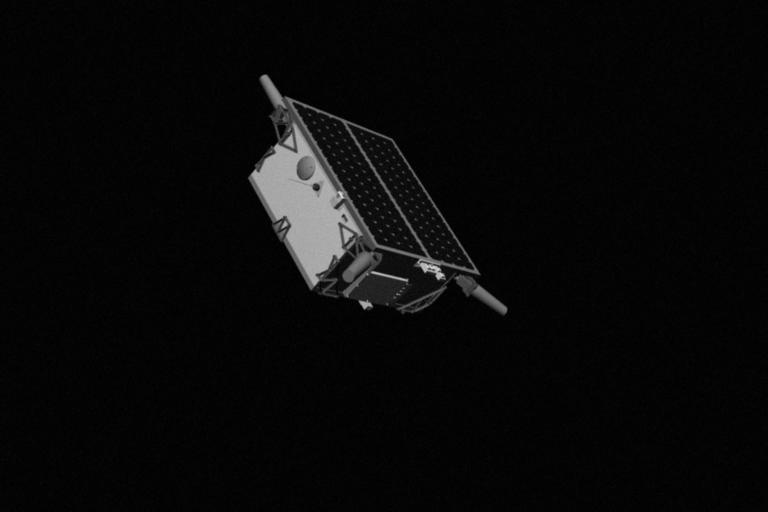}{1.6}{0.38}} &
    \Block{2-1}{\ImgER{0.205}{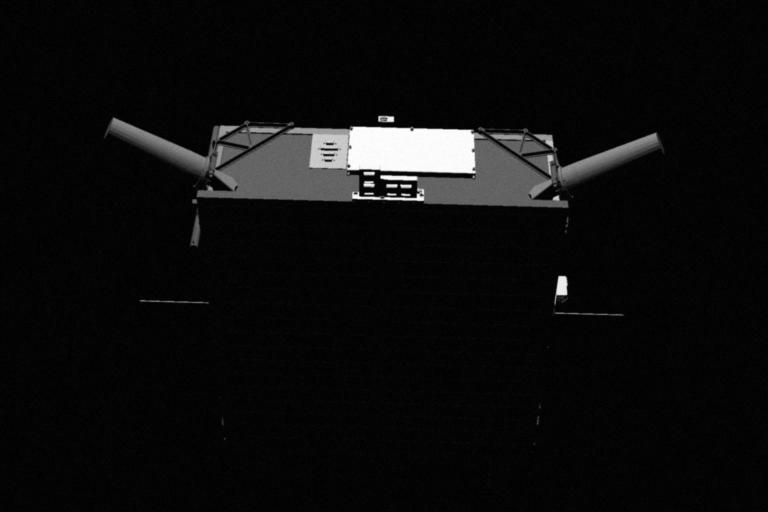}{2.1}{1.87}} &
    \Block{2-1}{\ImgER{0.205}{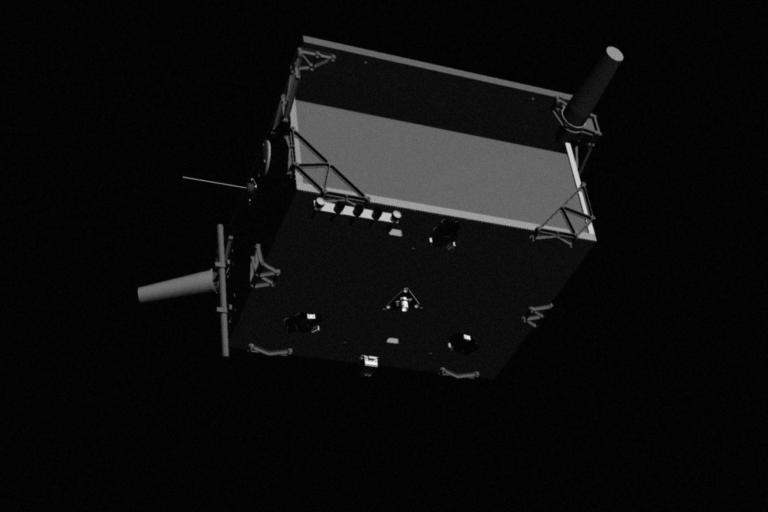}{2.1}{0.94}} &
    \Block{2-1}{\ImgER{0.205}{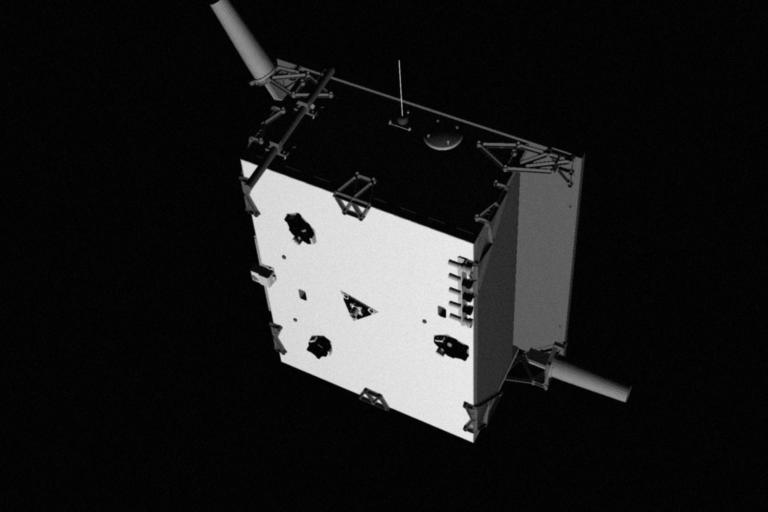}{2.5}{0.47}} \\
& & &  &  &  \\

\Block{4-1}{\rotatebox[origin=c]{90}{\small \ 3D cues used by $P_{\Theta}$  as learned by $G_{\Phi}$ }} & \Block{2-1}{\small$\mathcal{F}$ trained \\ with fixed \\ pos. enc.} & 
    \Block{2-1}{\ImgER{0.205}{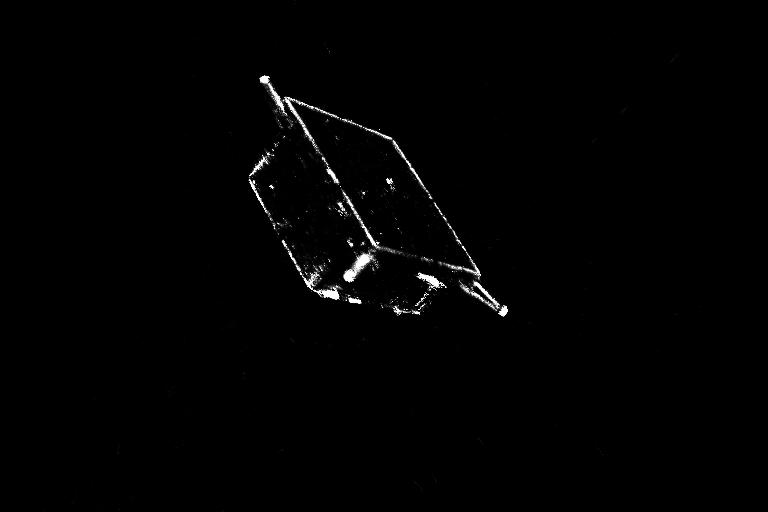}{1.9}{0.83}} &
    \Block{2-1}{\ImgER{0.205}{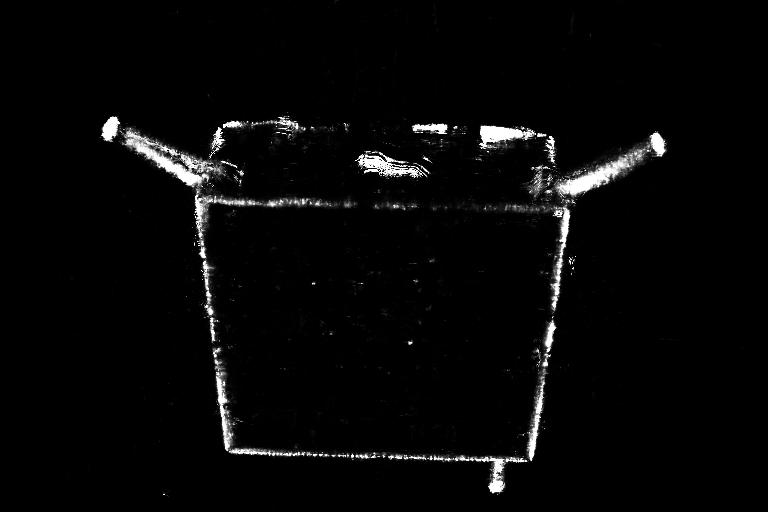}{12.5}{1.12}} &
    \Block{2-1}{\ImgER{0.205}{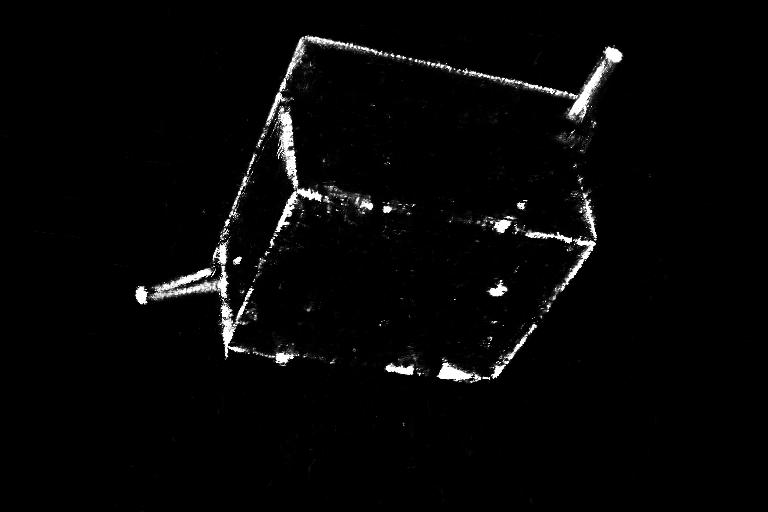}{10.9}{0.83}} &
    \Block{2-1}{\ImgER{0.205}{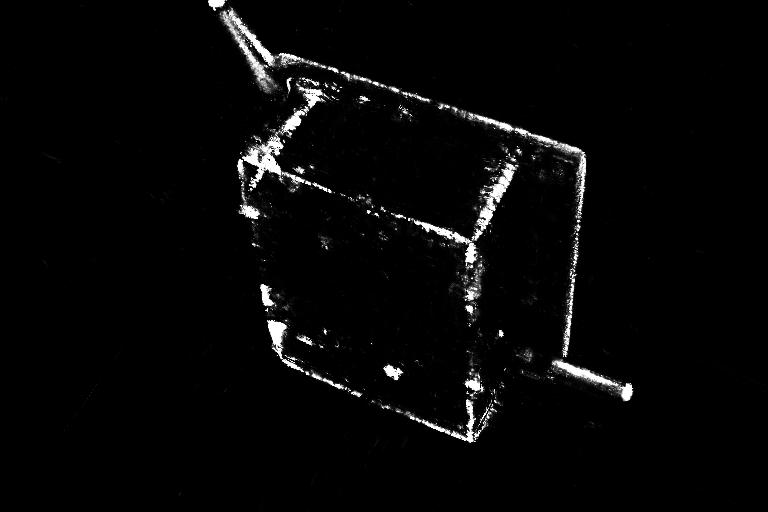}{11.4}{1.68}} \\

 &  &  &  & &  \\
 & \Block{2-1}{\small$\mathcal{F}$ \\ fully \\ trainable} & 
    \Block{2-1}{\ImgER{0.205}{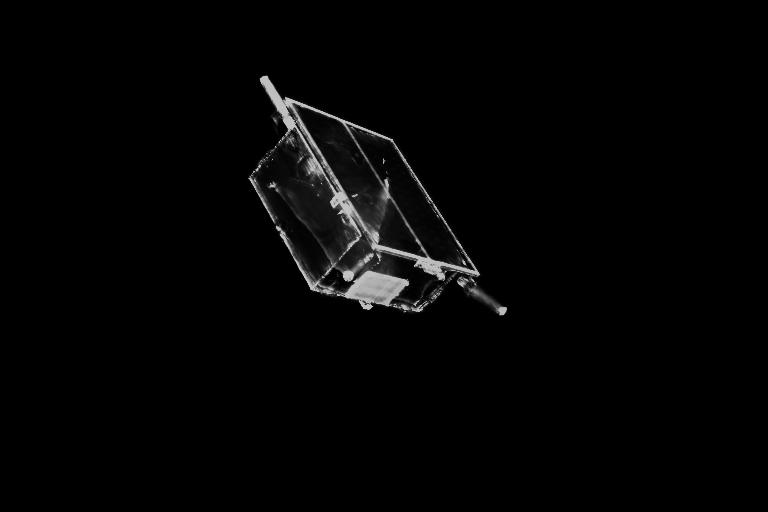}{2.8}{0.94}} &
    \Block{2-1}{\ImgER{0.205}{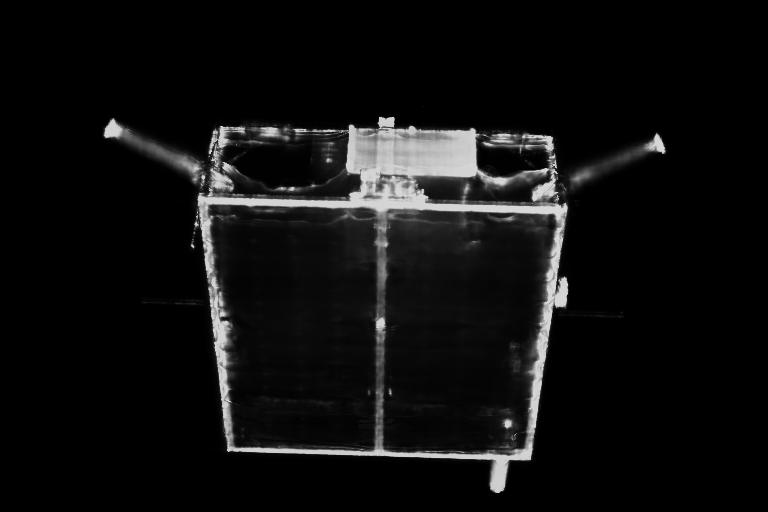}{1.9}{0.49}} &
    \Block{2-1}{\ImgER{0.205}{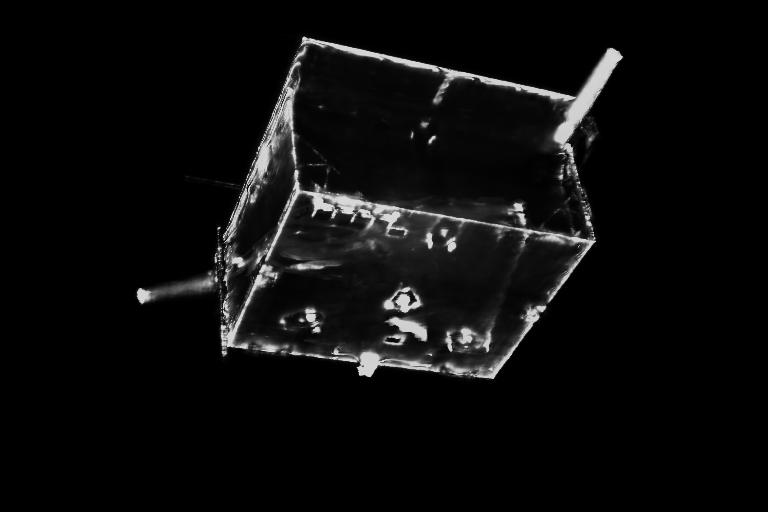}{4.8}{1.19}} &
    \Block{2-1}{\ImgER{0.205}{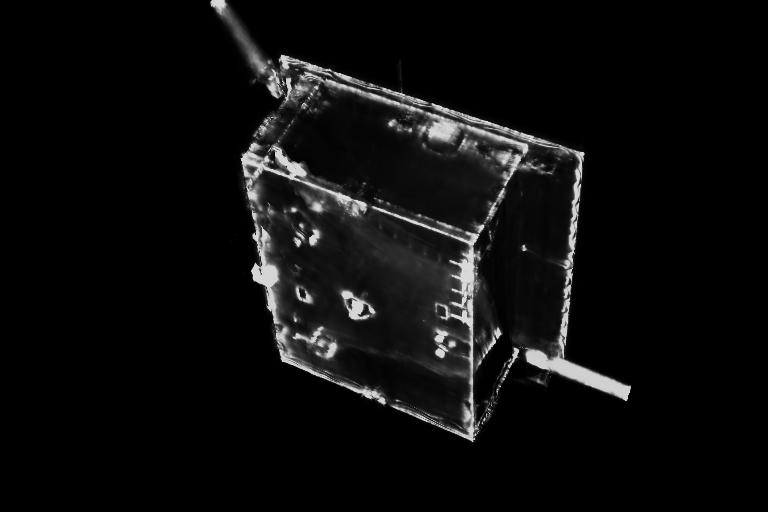}{4.2}{1.00}} \\
 
 &  &  &  &  &  \\
\end{NiceTabular}
\end{table*}

\subsection{Validation}
\label{sec_val_main}

In this section, we demonstrate that our visualization method is able to recover the key 3D features that are primarily exploited by a given spacecraft pose estimation network. For this purpose, we trained two image generation networks $G_{\Phi}$, based on the same NeRF architecture. Their only difference is the fact that the first network learns the whole neural field $\mathcal{F}$ (see \cref{fig_nerf_architecture}) while the second network uses a pretrained positional encoding which is frozen during training, \ie, only the density and color fields are learned. 

\Cref{tab_validation} depicts the visual cues learned by both networks for different viewpoints, along with the corresponding images from the SPEED+ dataset~\cite{park2022speed+}. Not only do both networks converge, but they also generate similar visual cues. Those mainly consist in structural edges and asymmetrical features such as antennas, sun sensors, or a torque rod. However, the second model, \ie, which uses a pretrained positional encoding, is able to recover smaller details. Furthermore, it only requires 1,000 optimization steps instead of the 3,000 steps required by the first one due to the regularization effect of the positional encoding. For these reasons, the next experiments are always conducted with an image generator for which the positional encoding was pretrained and is not updated during the optimization.

\Cref{tab_validation} also mentions the error metrics computed on the predictions made by the pose estimator on the rendered visual cues. It shows that the pose estimation network achieves a similar accuracy on the generated visual cues than on the synthetic images. Those visual cues can therefore be interpreted as the main target features on which the pose estimator relies.

\renewcommand{\arraystretch}{3.2}
\begin{table*}[t]
\vspace{0.5cm}
\centering
\caption{ \label{tab_supervision} \small 3D cues rendered by a NeRF $G_{\Phi}$ trained with our method on a pose estimator $P_{\Theta}$, considering two training configurations for both $G_{\Phi}$ and for $P_{\Theta}$. Training $G_{\Phi}$ through the pose loss encourages the learning of pose-relevant singularities (peripheral or asymmetrical), while the heatmap loss forces the reconstruction of edges. A pose estimation network trained only on a pose regression task relies on less robust and detailed features, which results in poorer generalization capabilities.}
\small
\begin{NiceTabular}{|@{}m{1.6cm}@{}|@{}m{1.9cm}@{}|@{}m{1.9cm}|@{}m{1.8cm}@{}*{3}{|@{}c@{}}|}[hlines]

\makecell{$P_{\Theta}$ \\ Training} & \makecell{\textit{Lightbox} \\ Metrics} & \makecell{\textit{Sunlamp} \\ Metrics} & \makecell{$G_{\Phi}$ \\ Supervision} & \Block{1-3}{Visual Cues} \\

\Block{4-1}{\makecell[c]{Heatmap \\ \& \\ Pose}} &
\Block{4-1}{\makecell[c]{$E_R$ = 9.4\degree \vspace{0.2cm} \\ $E_T$ = 24cm}} & 
\Block{4-1}{\makecell[c]{$E_r$ = 15.5\degree \vspace{0.2cm} \\$E_T$ = 28cm}} &
\Block{2-1}{\centering Heatmap} &
\Block{2-1}{\includegraphics[width=0.19\linewidth]{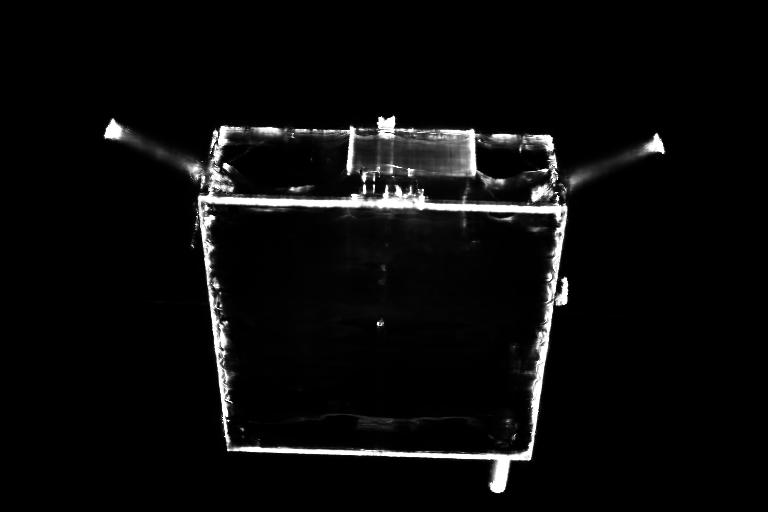}\vspace{-0.1cm}} &
\Block{2-1}{\includegraphics[width=0.19\linewidth]{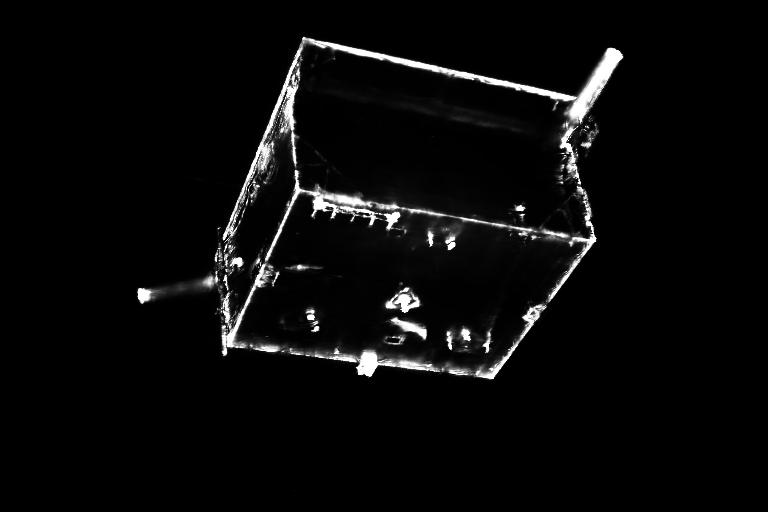}\vspace{-0.1cm}} & 
\Block{2-1}{\includegraphics[width=0.19\linewidth]{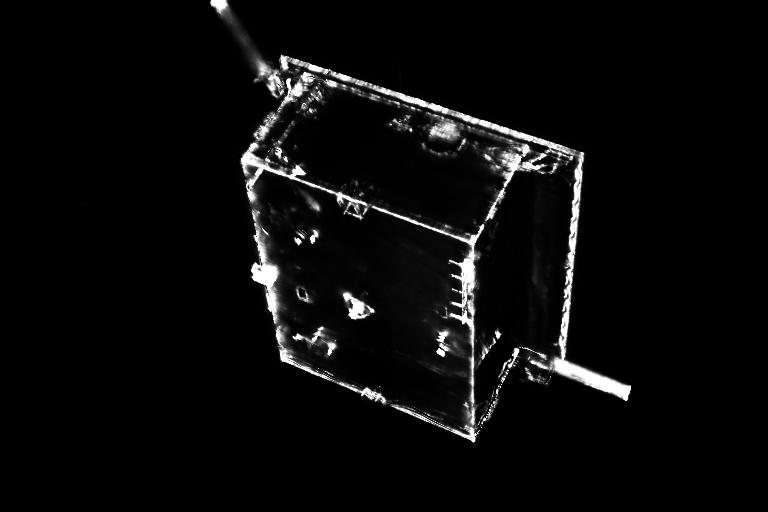}\vspace{-0.1cm}} \\

& & & & & & \\
& & & \Block{2-1}{\centering Pose} &
\Block{2-1}{\includegraphics[width=0.19\linewidth]{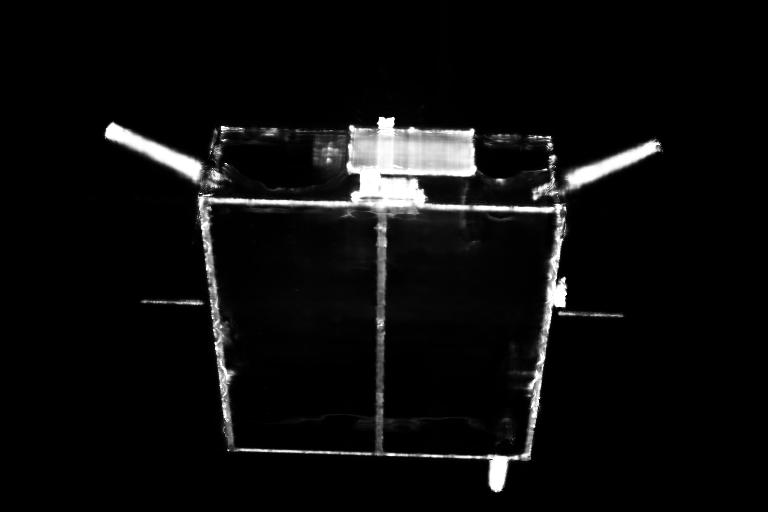}\vspace{-0.1cm}} &
\Block{2-1}{\includegraphics[width=0.19\linewidth]{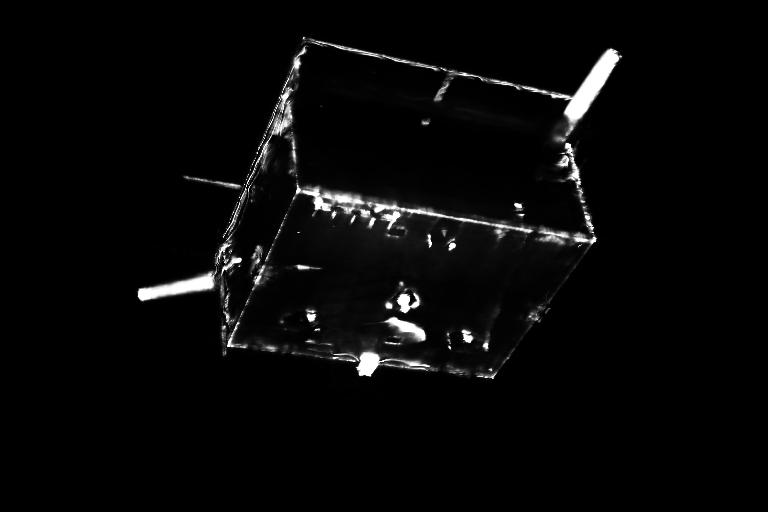}\vspace{-0.1cm}} &
\Block{2-1}{\includegraphics[width=0.19\linewidth]{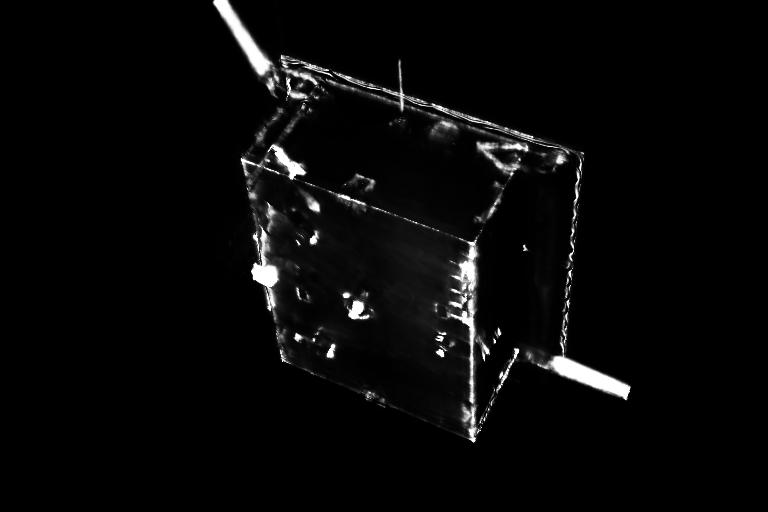}\vspace{-0.1cm}} \\
& & & & & & \\

\Block{2-1}{Pose} & \Block{2-1}{\makecell[c]{$E_R$ = 24.5\degree \vspace{0.2cm} \\ $E_T$ = 46 cm}} &  \Block{2-1}{\makecell[c]{ $E_R$ = 34.9\degree \vspace{0.2cm}  \\ $E_T$ = 38 cm}} & \Block{2-1}{Pose} &
\Block{2-1}{\includegraphics[width=0.19\linewidth]{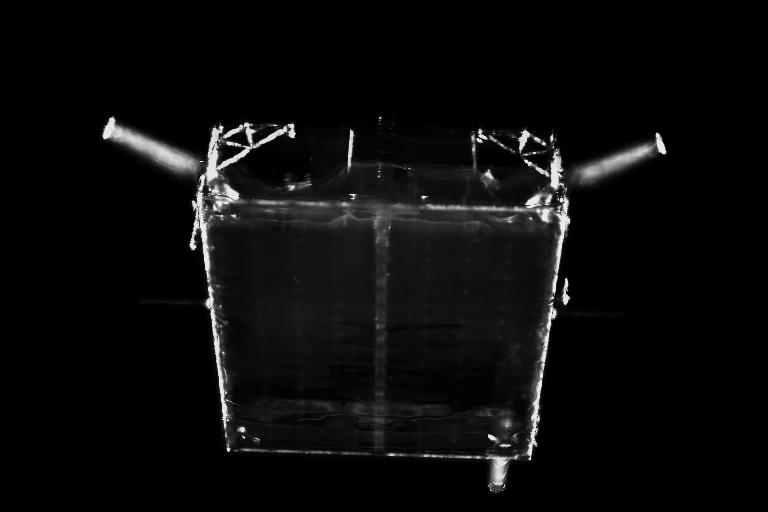}\vspace{-0.1cm}} &
\Block{2-1}{\includegraphics[width=0.19\linewidth]{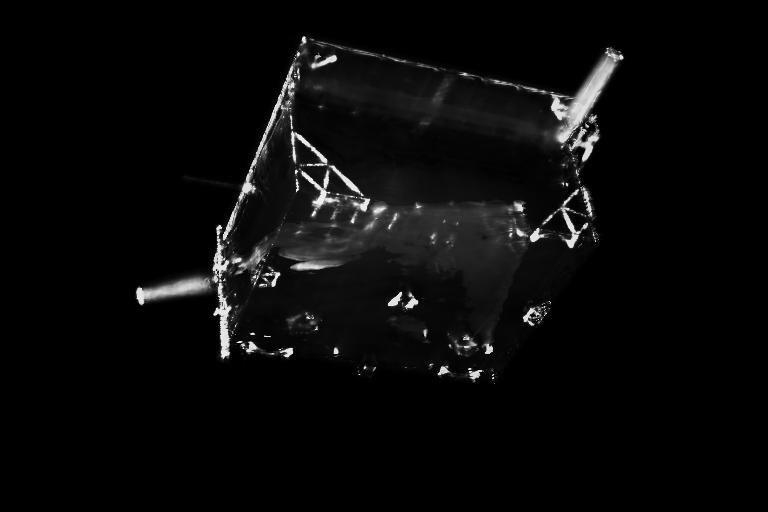}\vspace{-0.1cm}} &
\Block{2-1}{\includegraphics[width=0.19\linewidth]{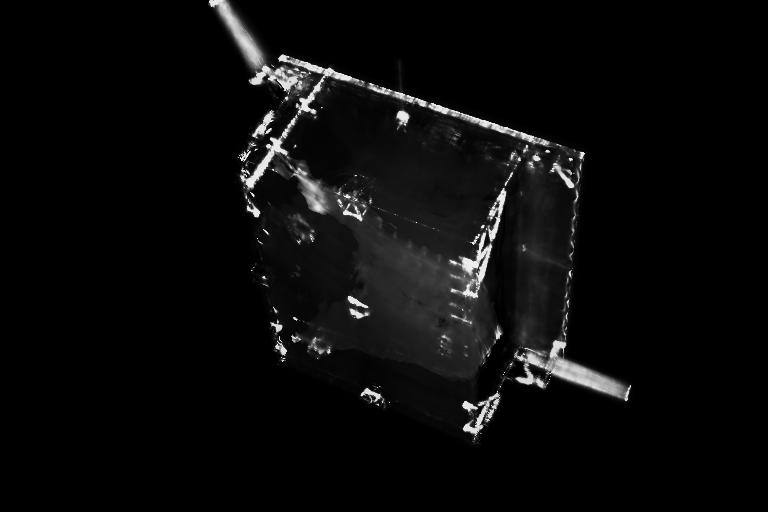}\vspace{-0.1cm}} \\
 
& & & & & & \\
\end{NiceTabular}
\end{table*}

\subsection{Insights on the Supervision of Pose Estimators}
\label{sec_val_insights}

    In the previous section, we considered an image generation network trained through the back-propagation of a loss that grasps the pose estimator ability of both detecting pre-defined keypoints and predicting the target relative pose. In this section, we consider two image generation networks learned through the back-propagation of either the heatmap-based keypoint regression loss or the pose regression one. 
    
    \Cref{tab_supervision} depicts the images rendered by these image generation models. Visual cues from both models share similarities with the ones obtained through the combined loss. Indeed, they show that the model mostly relies on edges for both tasks. However, a key difference between both visualization models lies in the details that the model trained on the pose loss recovers. Indeed, this model learns subtle asymmetrical features such as the two thin antennas on the spacecraft sides. Together with the three strongly highlighted main antennas, they form a set of pose-relevant singularities. These singularities are especially useful for the pose estimation task as they are asymmetrical features that are distant from the spacecraft center, thereby enabling more accurate pose predictions. In contrast, the heatmap-based keypoint regression task does not really benefit from those features so that the image generation network trained on this loss does not learn to reconstruct them. Instead, it mainly relies on edges that are resilient to photometric variations and are relevant to the position of keypoints.

    As we have seen, both tasks rely on similar features, except for the pose-relevant singularities exploited by the pose estimation head. An obvious question arises from this observation: "Is the double supervision, really required to train the pose estimation network ?" To answer to that question, we first trained a SPNv2 model with only a pose estimation head. Then, we trained an image generator using the pose estimation loss. \Cref{tab_supervision} depicts the visual cues learned by the image generation network.

    The pose-supervised 3D cues no longer retain the distinct pose-relevant features previously observed. The thin side antennas of the spacecraft are missing from the reconstruction, and the three primary antennas lack their earlier contrast. Additionally, artifacts resembling to some elements of the synthetic SPEED+ images have emerged, such as the solar panel texture. This indicates that the pose estimator trained only on the pose regression task has overfitted on the training set by learning a representation that is not as robust as the one learned through the pose estimator supervision of both tasks. This visual observation is consistent with the pose estimation accuracy metrics measured on the HIL sets of SPEED+ and reported in \cref{tab_supervision} which show that the network poorly generalizes. Through the visualization of the 3D target features on which the model relies, our method has therefore the potential of indicating to which extent a pose estimation network has learned robust features, or, conversely, has overfitted to the training set distribution.
    
\section{CONCLUSION}
    In this paper, we introduced a method for visualizing the 3D cues which steer the prediction of a given spacecraft pose estimation network. This visualization is of a particular interest for understanding the decision processes of spacecraft pose estimation networks and therefore aims at alleviating the "black-box" effect that hamper their adoption within the space robotics community. 
    
    By training an image generation network using the gradients back-propagated through the pose estimator, we enforce the image generator to recover the target 3D features that contribute the most to the pose estimator predictions. To alleviate the convergence issues related to a challenging problem formulation, the image generation network consists in a Neural Radiance Field, trained through gradient accumulation to ensure the 3D consistency of the learned features. 
    
    Experiments carried out with a state-of-the-art spacecraft pose estimation network, \ie, SPNv2, demonstrate that the proposed method is able to recover the 3D cues on which this pose estimation network relies. In addition, our method helps in gaining insights on the relationship between the pose estimation network supervision and the visual cues exploited by that estimator and demonstrated the interest of multi-task learning in improving the generalization capabilities of a pose estimation network.


\section*{ACKNOWLEDGMENT}
    Computational resources have been provided by the Consortium des Équipements de Calcul Intensif (CÉCI), funded by the Fonds de la Recherche Scientifique de Belgique (F.R.S.-FNRS) under Grant No. 2.5020.11 and by the Walloon Region.


\begin{thebibliography}{99}
\bibitem{henshaw2014darpa} C. G. Henshaw, The darpa phoenix spacecraft servicing program: Overview and plans for risk reduction, in \textit{i-SAIRAS}, European Space Agency, 2014.
\bibitem{biesbroek2021clearspace} R. Biesbroek, S. Aziz, A. Wolahan, S. Cipolla, M. Richard-Noca, and L. Piguet, The clearspace-1 mission: ESA and clearspace team up to remove debris, in \textit{Proc. 8th Eur. Conf. Sp. Debris}, pp. 1--3, 2021.
\bibitem{sharma2018pose} S. Sharma, C. Beierle, and S. D'Amico, Pose estimation for non-cooperative spacecraft rendezvous using convolutional neural networks, in \textit{2018 IEEE Aerospace Conference}, pp. 1--12, IEEE, 2018.
\bibitem{opromolla2017review} R. Opromolla, G. Fasano, G. Rufino, and M. Grassi, A review of cooperative and uncooperative spacecraft pose determination techniques for close-proximity operations, \textit{Progress in Aerospace Sciences}, vol. 93, pp. 53--72, 2017. doi:10.1016/j.paerosci.2017.07.001.
\bibitem{christian2013survey} J. A. Christian and S. Cryan, A survey of LIDAR technology and its use in spacecraft relative navigation, in \textit{AIAA Guidance, Navigation, and Control (GNC) Conference}, pp. 4641, 2013. doi:10.2514/6.2013-4641.
\bibitem{shi2015uncooperative} J.-F. Shi, S. Ulrich, S. Ruel, and M. Anctil, Uncooperative spacecraft pose estimation using an infrared camera during proximity operations, in \textit{AIAA SPACE 2015 conference and exposition}, pp. 4429, 2015. doi:10.2514/6.2015-4429.
\bibitem{rondao2022chinet} D. Rondao, N. Aouf, and M. A. Richardson, ChiNet: Deep recurrent convolutional learning for multimodal spacecraft pose estimation, \textit{IEEE Transactions on Aerospace and Electronic Systems}, vol. 59, no. 2, pp. 937--949, 2022. doi:10.1109/TAES.2022.3193085.
\bibitem{martinez2017pose} H. G. Martínez, G. Giorgi, and B. Eissfeller, Pose estimation and tracking of non-cooperative rocket bodies using time-of-flight cameras, \textit{Acta Astronautica}, vol. 139, pp. 165--175, 2017. doi:10.1016/j.actaastro.2017.07.002.
\bibitem{pesce2017stereovision} V. Pesce, M. Lavagna, and R. Bevilacqua, Stereovision-based pose and inertia estimation of unknown and uncooperative space objects, \textit{Advances in Space Research}, vol. 59, no. 1, pp. 236--251, 2017. doi:10.1016/j.asr.2016.10.002.
\bibitem{kisantal2020satellite} M. Kisantal, S. Sharma, T. H. Park, D. Izzo, M. Märtens, and S. D’Amico, Satellite pose estimation challenge: Dataset, competition design, and results, \textit{IEEE Transactions on Aerospace and Electronic Systems}, vol. 56, no. 5, pp. 4083--4098, 2020.
\bibitem{pauly2023survey} L. Pauly, W. Rharbaoui, C. Shneider, A. Rathinam, V. Gaudilliere, and D. Aouada, A survey on deep learning-based monocular spacecraft pose estimation: Current state, limitations and prospects, \textit{Acta Astronautica}, vol. 212, pp. 339--360, 2023.
\bibitem{park2023satellite} T. H. Park, M. Märtens, M. Jawaid, Z. Wang, B. Chen, T.-J. Chin, D. Izzo, and S. D’Amico, Satellite pose estimation competition 2021: Results and analyses, \textit{Acta Astronautica}, vol. 204, pp. 640--665, 2023.
\bibitem{cosmas2020fpga} K. Cosmas and K. Asami, Utilization of FPGA for onboard inference of landmark localization in CNN-Based spacecraft pose estimation, \textit{Aerospace}, vol. 7, no. 11, pp. 159, 2020. doi:10.3390/aerospace7110159.
\bibitem{posso2024real} J. Posso, G. Bois, and Y. Savaria, Real-Time Spacecraft Pose Estimation Using Mixed-Precision Quantized Neural Network on COTS Reconfigurable MPSoC, in \textit{2024 22nd IEEE Interregional NEWCAS Conference (NEWCAS)}, pp. 358--362, IEEE, 2024.
\bibitem{chen2019satellite} B. Chen, J. Cao, A. Parra, and T.-J. Chin, Satellite pose estimation with deep landmark regression and nonlinear pose refinement, in \textit{Proceedings of the IEEE/CVF international conference on computer vision workshops}, pp. 0--0, 2019.
\bibitem{cassinis2022ground} L. P. Cassinis, A. Menicucci, E. Gill, I. Ahrns, and M. Sanchez-Gestido, On-ground validation of a CNN-based monocular pose estimation system for uncooperative spacecraft: Bridging domain shift in rendezvous scenarios, \textit{Acta Astronautica}, vol. 196, pp. 123--138, 2022.
\bibitem{lepetit2009ep} V. Lepetit, F. Moreno-Noguer, and P. Fua, EP n P: An accurate O (n) solution to the P n P problem, \textit{International journal of computer vision}, vol. 81, pp. 155--166, 2009.
\bibitem{proencca2020deep} P. F. Proença and Y. Gao, Deep learning for spacecraft pose estimation from photorealistic rendering, in \textit{2020 IEEE International Conference on Robotics and Automation (ICRA)}, pp. 6007--6013, IEEE, 2020.
\bibitem{posso2022mobile} J. Posso, G. Bois, and Y. Savaria, Mobile-ursonet: an embeddable neural network for onboard spacecraft pose estimation, in \textit{2022 IEEE International Symposium on Circuits and Systems (ISCAS)}, pp. 794--798, IEEE, 2022.
\bibitem{legrand2022end} A. Legrand, R. Detry, and C. De Vleeschouwer, End-to-end neural estimation of spacecraft pose with intermediate detection of keypoints, in \textit{European Conference on Computer Vision}, pp. 154--169, Springer, 2022.
\bibitem{park2024robust} T. H. Park and S. D’Amico, Robust multi-task learning and online refinement for spacecraft pose estimation across domain gap, \textit{Advances in Space Research}, vol. 73, no. 11, pp. 5726--5740, 2024.
\bibitem{legrand2024domain} A. Legrand, R. Detry, and C. De Vleeschouwer, Domain Generalization for In-Orbit 6D Pose Estimation, in \textit{Journal of Aerospace Information Systems}, pp 1--10, 2025.
\bibitem{selvaraju2017grad} R. R. Selvaraju, M. Cogswell, A. Das, R. Vedantam, D. Parikh, and D. Batra, Grad-cam: Visual explanations from deep networks via gradient-based localization, in \textit{Proceedings of the IEEE International Conference on Computer Vision}, pp. 618--626, 2017.
\bibitem{smilkov2017smoothgrad} D. Smilkov, N. Thorat, B. Kim, F. Viégas, and M. Wattenberg, Smoothgrad: removing noise by adding noise, \textit{arXiv preprint arXiv:1706.03825}, 2017.
\bibitem{sundararajan2017axiomatic} M. Sundararajan, A. Taly, and Q. Yan, Axiomatic attribution for deep networks, in \textit{International Conference on Machine Learning}, pp. 3319--3328, PMLR, 2017.
\bibitem{ribeiro2016should} M. T. Ribeiro, S. Singh, and C. Guestrin, "Why should I trust you?" Explaining the predictions of any classifier, in \textit{Proceedings of the 22nd ACM SIGKDD International Conference on Knowledge Discovery and Data Mining}, pp. 1135--1144, 2016.
\bibitem{petsiuk2018rise} V. Petsiuk, A. Das, and K. Saenko, Rise: Randomized input sampling for explanation of black-box models, \textit{arXiv preprint arXiv:1806.07421}, 2018.
\bibitem{englebert2024poly} A. Englebert, O. Cornu, and C. De Vleeschouwer, Poly-cam: high resolution class activation map for convolutional neural networks, \textit{Machine Vision and Applications}, vol. 35, no. 4, pp. 89, Springer, 2024.
\bibitem{englebert2022backward} A. Englebert, O. Cornu, and C. De Vleeschouwer, Backward recursive class activation map refinement for high resolution saliency map, in \textit{2022 26th International Conference on Pattern Recognition (ICPR)}, pp. 2444--2450, IEEE, 2022.
\bibitem{mildenhall2021nerf} B. Mildenhall, P. P. Srinivasan, M. Tancik, J. T. Barron, R. Ramamoorthi, and R. Ng, Nerf: Representing scenes as neural radiance fields for view synthesis, \textit{Communications of the ACM}, vol. 65, no. 1, pp. 99--106, ACM New York, NY, USA, 2021.
\bibitem{legrand2024leveraging} A. Legrand, R. Detry, and C. De Vleeschouwer, Leveraging Neural Radiance Fields for Pose Estimation of an Unknown Space Object during Proximity Operations, \textit{2024 IEEE ISpaRo, available at arXiv:2405.12728}, 2024.
\bibitem{legrand2024domain_nerf} A. Legrand, R. Detry, and C. De Vleeschouwer, Domain generalization for 6D pose estimation through NeRF-based image synthesis, \textit{arXiv preprint arXiv:2407.10762}, 2024.
\bibitem{fridovich2023k} S. Fridovich-Keil, G. Meanti, F. R. Warburg, B. Recht, and A. Kanazawa, K-planes: Explicit radiance fields in space, time, and appearance, in \textit{Proceedings of the IEEE/CVF Conference on Computer Vision and Pattern Recognition}, pp. 12479--12488, 2023.
\bibitem{ma2017pose} L. Ma, X. Jia, Q. Sun, B. Schiele, T. Tuytelaars, and L. Van Gool, Pose guided person image generation, \textit{Advances in Neural Information Processing Systems}, vol. 30, 2017.
\bibitem{nguyen2019hologan} T. Nguyen-Phuoc, C. Li, L. Theis, C. Richardt, and Y.-L. Yang, Hologan: Unsupervised learning of 3d representations from natural images, in \textit{Proceedings of the IEEE/CVF International Conference on Computer Vision}, pp. 7588--7597, 2019.
\bibitem{schwarz2020graf} K. Schwarz, Y. Liao, M. Niemeyer, and A. Geiger, Graf: Generative radiance fields for 3d-aware image synthesis, \textit{Advances in Neural Information Processing Systems}, vol. 33, pp. 20154--20166, 2020.
\bibitem{chan2021pi} E. R. Chan, M. Monteiro, P. Kellnhofer, J. Wu, and G. Wetzstein, pi-gan: Periodic implicit generative adversarial networks for 3d-aware image synthesis, in \textit{Proceedings of the IEEE/CVF Conference on Computer Vision and Pattern Recognition}, pp. 5799--5809, 2021.
\bibitem{chan2022efficient} E. R. Chan, C. Z. Lin, M. A. Chan, K. Nagano, B. Pan, S. De Mello, O. Gallo, L. J. Guibas, J. Tremblay, S. Khamis, et al., Efficient geometry-aware 3d generative adversarial networks, in \textit{Proceedings of the IEEE/CVF Conference on Computer Vision and Pattern Recognition}, pp. 16123--16133, 2022.
\bibitem{muller2022instant} T. Müller, A. Evans, C. Schied, and A. Keller, Instant neural graphics primitives with a multiresolution hash encoding, \textit{ACM Transactions on Graphics (TOG)}, vol. 41, no. 4, pp. 1--15, ACM New York, NY, USA, 2022.
\bibitem{park2022speed+} T. H. Park, M. Märtens, G. Lecuyer, D. Izzo, and S. D'Amico, SPEED+: Next-generation dataset for spacecraft pose estimation across domain gap, in \textit{2022 IEEE Aerospace Conference (AERO)}, pp. 1--15, IEEE, 2022.
\bibitem{gill2007autonomous} E. Gill, S. D’Amico, and O. Montenbruck, Autonomous formation flying for the PRISMA mission, \textit{Journal of Spacecraft and Rockets}, vol. 44, no. 3, pp. 671--681, 2007.
\bibitem{tan2019efficientnet} M. Tan and Q. Le, Efficientnet: Rethinking model scaling for convolutional neural networks, in \textit{International Conference on Machine Learning}, pp. 6105--6114, PMLR, 2019.
\bibitem{bukschat2020efficientpose} Y. Bukschat and M. Vetter, EfficientPose: An efficient, accurate and scalable end-to-end 6D multi object pose estimation approach, \textit{arXiv preprint arXiv:2011.}

\end{thebibliography}

\addtolength{\textheight}{-12cm}   

\end{document}